\documentclass[runningheads]{llncs}
\usepackage{graphicx}
\usepackage{amssymb}
\usepackage{algorithm}
\usepackage{algorithmic}
\usepackage[misc]{ifsym} 

\begin{document}
\makeatletter
\renewcommand*{\@fnsymbol}[1]{\ensuremath{\ifcase#1\or *\or \dagger\or \ddagger\or
   \mathsection\or \mathparagraph\or \|\or **\or \dagger\dagger
   \or \ddagger\ddagger \else\@ctrerr\fi}}
\makeatother

\title{Enabling surrogate-assisted evolutionary reinforcement learning via policy embedding\thanks{This work was supported partly by the National Natural Science Foundation of China (Grants 62272210 and 62106099), partly by the Guangdong Provincial Key Laboratory (Grant 2020B121201001), partly by the Program for Guangdong Introducing Innovative and Entrepreneurial Teams (Grant 2017ZT07X386), and partly by the Stable Support Plan Program of Shenzhen Natural Science Fund (Grant 20200925154942002).}}
%
\author{
Lan Tang\inst{1}
\and
Xiaxi Li\inst{2}\thanks{Lan Tang and Xiaxi Li contributed equally to this work.}
\and
Jinyuan Zhang\inst{1}
\and
Guiying Li\inst{1,3} 
\and
Peng Yang\inst{1,4}\thanks{Corresponding author: Peng Yang
\\E-mail address: yangp@sustech.edu.cn}
(\Letter)
\and
Ke Tang\inst{1,3}
}

\authorrunning{L. Tang and X. Li et al.}
\titlerunning{Enabling surrogate-assisted ERL via policy embedding}

%

\institute{
Guangdong Provincial Key Laboratory of Brain-Inspired Intelligent Computation, Department of Computer Science and Engineering, Southern University of Science and Technology, Shenzhen 518055, China 
\and
Faculty of engineering, Shenzhen MSU-BIT university, Shenzhen 518172, China
\and
Research Institute of Trustworthy Autonomous Systems, Southern University of Science and Technology, Shenzhen 518055, China
\and
Department of Statistics and Data Science, Southern University of Science and Technology, Shenzhen 518055, China
}
\maketitle              

\begin{abstract}
Evolutionary Reinforcement Learning (ERL) that applying Evolutionary Algorithms (EAs) to optimize the weight parameters of Deep Neural Network (DNN) based policies has been widely regarded as an alternative to traditional reinforcement learning methods. However, the evaluation of the iteratively generated population usually requires a large amount of computational time and can be prohibitively expensive, which may potentially restrict the applicability of ERL. Surrogate is often used to reduce the computational burden of evaluation in EAs. Unfortunately, in ERL, each individual of policy usually represents millions of weights parameters of DNN. This high-dimensional representation of policy has introduced a great challenge to the application of surrogates into ERL to speed up training. This paper proposes a PE-SAERL Framework to at the first time enable surrogate-assisted evolutionary reinforcement learning via policy embedding (PE). 
Empirical results on 5 Atari games show that the proposed method can perform more efficiently than the four state-of-the-art algorithms.
The training process is accelerated up to 7x on tested games, comparing to its counterpart without the surrogate and PE.

\end{abstract}
%
%
%
%
%
%

\section{Introduction}
\label{sect:introduction}

In recent years, Evolutionary Algorithms (EAs) have been successfully applied to exploratively optimize the parameters of the Deep Neural Network (DNN) based policy for challenging reinforcement learning tasks, i.e., video games \cite{OpenAI_ES,CCNCS}, Robotics \cite{robot}, and computer vision \cite{ERL_CV}.
The resultant Evolutionary Reinforcement Learning (ERL) \cite{review_ERL} has been widely regarded as an alternative to traditional reinforcement learning methods \cite{OpenAI_ES},  due to the ability of EAs on exploration, noise resistance, and parallel acceleration. 
In ERL, a diverse population of policies is generated by EA and is evaluated via interactions with the environment. The fitness value of each individual is considered as the total returns (e.g., cumulative rewards with discount factor 1.0) across the entire episode. 

However, there is a notable issue with ERL \cite{review_ERL} : the evaluation of the population usually requires a large amount of computational time and can be prohibitively expensive. 
First, ERL finds the optimal policy by iteratively searching in a population-based manner. This manner leads to the number of evaluation times of individual policies being numerous. 
Second, each real fitness evaluation can be computationally time-consuming due to the complex mechanism of the reinforcement learning simulators. 
These unfavorable characteristics can potentially restrict the application of ERL in real-world problems.

Actually, the computationally expensive problems have been extensively studied in the EA community for decades. Many Surrogate-Assisted Evolutionary Algorithms (SAEAs) have been proposed for expensive problems. 
In typical SAEAs, surrogate models are trained to replace the expensive fitness function for evaluation \cite{SAEA_review_1}.
Based on this idea, this paper focus on how to apply surrogates into ERL to speed up training while keeping the ERL effective. 
And we call this framework as Surrogate-Assisted ERL (SAERL).
In this paper, we argue that the existing SAEA approaches can hardly be directly applied to ERL. 
The reason is that each individual in ERL is usually a Deep Neural Network (DNN) based policy with millions of weights. 
This requires the surrogates accepting very high dimensional input vectors to represent the parameters of individual policies.
This leads to three technical issues:
First, the surrogate model can be complex and computationally expensive. 
Second, the training of the surrogate model can be very difficult due to the lack of specific training data. 
Third, the surrogate model may encounter the difficulty of distinguishing different weight vectors of policies in high-dimensional space. 
Therefore, the existing surrogate approaches may fail to accelerate the training of ERL. Few exceptions \cite{ESP,SMB_NE,SA_ERL} employ the surrogates in the training phases, but they are not for high-dimensional DNN-based policies.

\begin{figure}[ht]
\centering
\includegraphics[width=0.9\columnwidth]{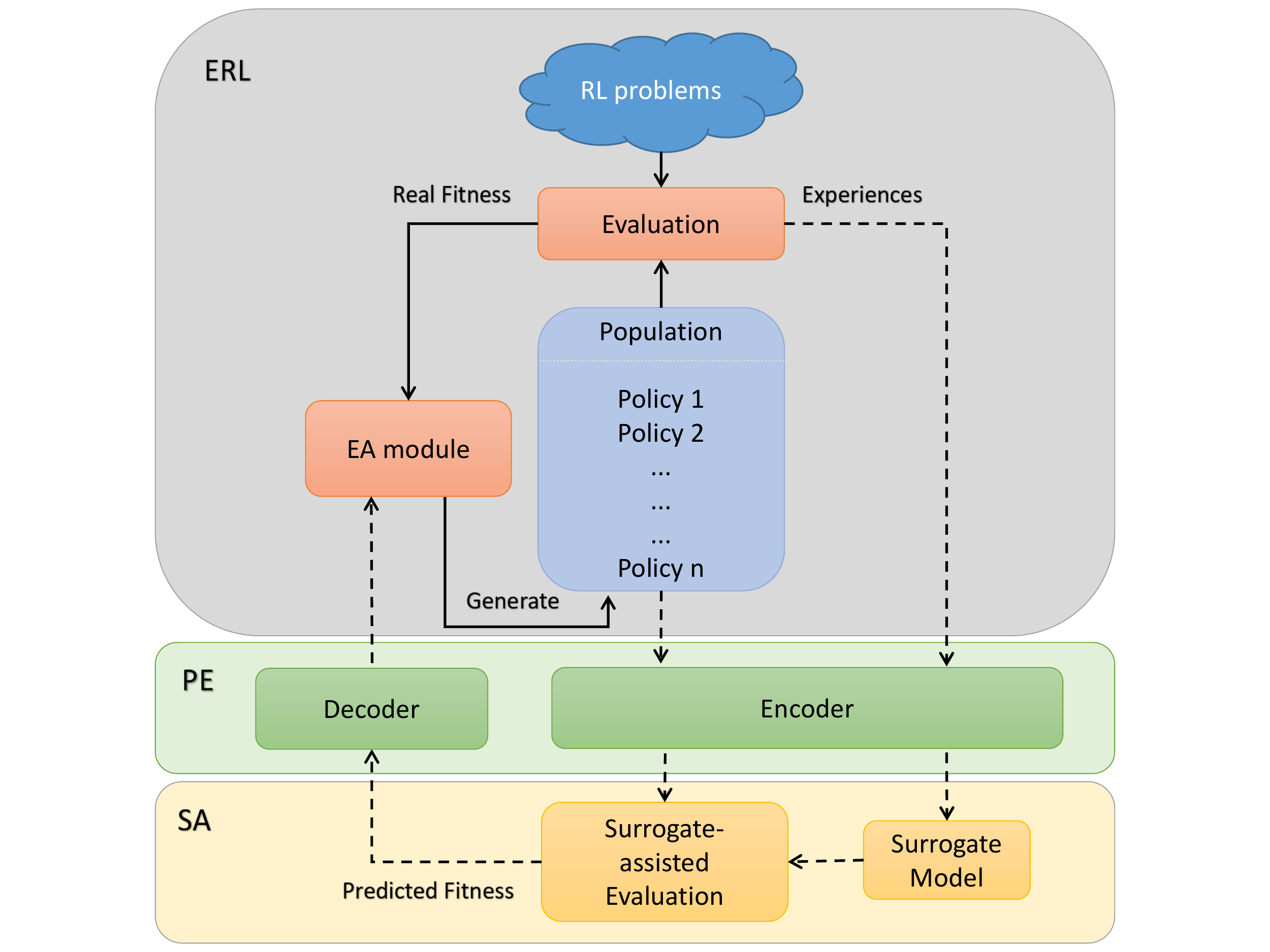} 
\caption{The overview of the PE-SAERL framework.}
\label{fig1}
\end{figure}

To eliminate the above issues, we first propose to project the surrogate model to work in a low dimensional space. A policy embedding (PE) module is proposed for the projection, and a new framework PE-SAERL is proposed. 
Specifically, before a high-dimensional policy vector is input into the surrogate model, the PE module first encodes the vector to a lower-dimensional space by preserving its major features. 
The existing surrogate models are thus able to be applied to the low-dimensional embedded input vectors for computationally cheap evaluation.
After the surrogate evaluation, the embedded policies are decoded to the original high-dimensional space for further processing of ERL, as shown in Fig. \ref{fig1}.

The PE module is essentially a bidirectional mapper of both high-dimensional and low-dimensional spaces. 
Also note that, the focus of this paper is the first time discussing how to technically enable the SAERL.
Hence, we are not focusing on sophisticated mappper designing. We restrict the complexity of the mapper used in this paper for better demonstrating that the PE module is useful for the SAERL framework.

Specifically, we choose the random embedding method \cite{RE_1} as the PE module here. 
This PE basically maps between the high-dimensional space and some randomly selected lower dimensions while is able to preserving the major features of the original space \cite{RE_1}. 
We believe other learning-based PE module can further improve the effectiveness of PE-SAERL, and this technically simple PE of random embedding can be viewed as an appropriate baseline for further investigations. 

To instantiated the PE-SAERL for empirical studies, the EA and the surrogate model should be specified. 
In this work, we choose Negatively Correlated Search (NCS) \cite{NCS} as the EA module and the Fuzzy Classification Pre-Selection (FCPS) \cite{zhang_FCPS} module as surrogate module. 
We should emphasize that generally any EA and its surrogate-assisted version can be adopted in the PE-SAERL framework for optimizing policies. 
Trade-offs will be considered between the computing complexity and the search effectiveness. 
As a result, we get an instantiated method called PE-FCPS-NCS. 
The empirical studies on 5 Atari games successfully verify that ERL with surrogate acceleration is highly competitive to traditional DRLs.

The remainder of this paper is organized as follows.
Section 2 reviews the related works.
Section 3 details the proposed method.
Section 4 presents the empirical studies on Atari games.
Finally, the conclusion will be drawn in Section 5.


\section{Related Works}
\label{sect:related_work}

This section briefly reviews the literature in the fields of surrogate-assisted evolutionary algorithms and evolutionary reinforcement learning.

\subsection{Surrogate-Assisted Evolutionary Algorithms}
\label{sect:SAEA}

Many surrogate modelling approaches have been introduced to SAEAs over the past few years.
According to whether the fitness function values of candidate solutions in the optimization process is directly predicted, it can be generally divided into two categories : absolute fitness models and relative fitness models \cite{SAEA_review_1}.


Absolute fitness models are commonly used in SAEAs \cite{SAEA_review_1}. They aim at predicting new candidate solutions’ fitness values by approximating the real fitness function, including Polynomial Regression (PR), Gaussian Process regression (a.k.a. Kriging model), Support Vector Regression (SVR), Radial Basis Function (RBF), Neural Network(NN) etc. 
Kriging \cite{wang_kriging,SMB_NE} is a popular surrogate model in SAEAs because it can estimate the uncertainty as well as the fitness.
SVR \cite{SVR} is an effective surrogate model which has some practical applications, such as the optimization of railway wind barriers, but its training process will be expensive when the dimensionality of the problem is very high. 
RBF models \cite{SAEA_six_comparison} have also been used in SAEAs to solve many real-world problems.



Relative fitness models focus on providing the relative rank or preference of candidate solutions rather than their absolute fitness values \cite{SAEA_review_1}.
Wang et al. in \cite{wang_SA_classification} proposed a hierarchical clustering algorithm by grouping the patient data into different numbers of clusters to reduce the amount of computation time. 
Subsequently, Wang et al. in \cite{wang_SA_1} employed an artificial neural network as a classifier to predict the dominance relationship between candidate solutions and reference solutions.
Zhou et al. in \cite{zhang_SA_fclustering} used a fuzzy clustering method to extract the weight vectors, which works by constructing new objectives as linear combinations of the original Many-objective optimization problems.
In addition, Zhang et al. in \cite{zhang_FCPS,zhang_SA_fc_boosting} applied a Fuzzy-KNN method to filter out 'unpromising' individuals before the actual evaluation.



\subsection{Evolutionary reinforcement learning}
\label{sect:ERL}

ERL has been developed for years \cite{ERL_book}, and we noticed that ERL attracts increasing attentions recently with the DNN-based policies. 
As early as 1999, David et al. in \cite{ERL_book} pointed out in the literature that evolutionary algorithm is a class of classical algorithms that search the space of policies for solving reinforcement learning tasks. 
Salimans et al. in \cite{OpenAI_ES} from OpenAI proposed to use a simplified natural evolution strategies to directly optimize the weights of policy neural networks, where the natural gradients was used to update a parameterized search distribution  in the direction of higher expected fitness. 
It proved that Evolutionary Algorithm (EA) is competitive for policy neural networks search in deep RL \cite{OpenAI_ES}. 
Subsequently, Chrabaszcz et al. in \cite{ERL_CES} compared a simper and basic canonical ES algorithm with OpenAI ES further confirmes that the power of ES-based algorithms for policy search may rival that of the gradient-based algorithms.
Recently, Yang et al. \cite{CCNCS} proposed the CCNCS method, which uses a novel search method called Negatively Correlated Search \cite{NCS,Peng_NCS_1,NCNES} by explicitly measuring the diversity among different search processes to drive them to the regions with uncertainty. Another related work is a hybrid algorithm \cite{ERL_2} combining gradient-based and derivative-free optimization.


Related to using surrogates in ERL, Wang et al. in \cite{SA_ERL} proposed Surrogate-assisted Controller (SC) applied on it to alleviate the computational burden of evaluation. Nevertheless, the core of SC is to replace the critic model, which aims to evaluate the action not the policy. In this regard, it does not follow the framework of SAEA and thus is quite different from the SAERL framework discussed in this work.
In addition, there exists a few studies on applying surrogates to enhance EAs on RL tasks, such as Evolutionary Surrogate-assisted Prescription (ESP) \cite{ESP}, surrogate model-based Neuroevolution (SMB-NE) \cite{SMB_NE}. However, these methods do not take the DNN into acount, and does not face the high-dimensional difficulties as is in this work.  



As can be seen from the literature, ERL is a extremely promising method for solving RL tasks, but the evaluation process is prohibitively expensive. 
Using the surrogate model can solve the problems of low sampling efficiency and expensive evaluation in traditional SAEAs.
However, considering that each individual in ERL usually involves millions of parameters and the commonly surrogate models (i.e., Kriging \cite{wang_kriging}, ${k}$NN \cite{zhang_SA_fc_boosting}) by calculating the genotype distance may have no way to extract effective features, the direct application of surrogates into ERL to speed up training may face inevitably failure. This article extends the surrogate model to ERL via policy embedding to speed up training.


\section{Methodology}
\label{sect:Methodology}

This section first introduces the proposed framework consisting of three main modules. Then we briefly describe random embedding employed by the PE module. Finally a concrete algorithm called PE-FCPS-NCS is described in detail.

\subsection{PE-SAERL Framework}
\label{sect:PE-SAERL}

Intuitively, we intend to apply surrogates into ERL to speed up training by replacing a part of the real simulations with more computationally cheap surrogates. Importantly, the PE module is designed to address the three key technical issues faced by directly applying the surrogates to millions of dimensions, as mentioned previously. PE-SAERL Framework fully follows the PE-SAERL flowchart depicted in Fig. \ref{fig1}.  


Specifically, the PE-SAERL Framework is roughly divided into 3 modules, which are ERL module, PE module and SA module.

\begin{figure}[htb]
\centering
\includegraphics[width=0.82\columnwidth]{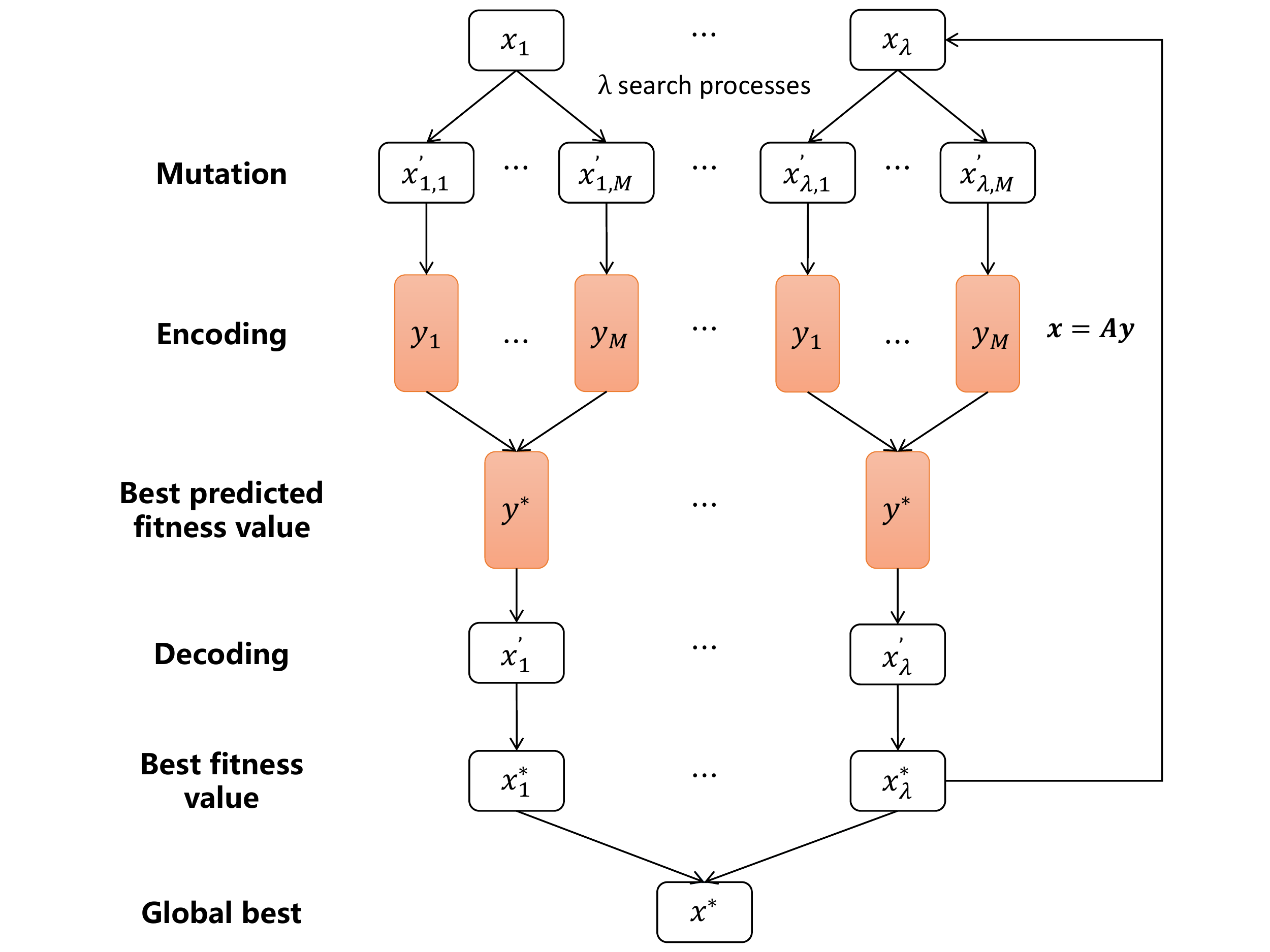} 
\caption{The parallel architecture and main processes of the PE-SAERL framework.}
\label{fig2}
\end{figure}

\begin{itemize}
\item[$\bullet$] \emph{ERL module}: ERL iteratively search for the optimal policy in a population-based manner with EAs. 
Each individual in the population is represented as a vector of all connection weights of the neural network policy. At the beginning of training, ${\lambda}$ search processes are initialized  in parallel. At each iteration, each of them separately preserves a individual ${x_i}$, and generates an offspring individual ${x_i^{{\prime}}}$ by crossover or variation operators. Eventually, under the effect of environmental selection, the individual with higher fitness value as new parent individual into the next generation, As depicted in Fig. \ref{fig2}.

\item[$\bullet$] \emph{PE module}: The PE module is essentially a bidirectional mapper of both high-dimensional and low-dimensional spaces. It generally consists of two steps, i.e., encoding and decoding. Encoding extracts a low-dimensional vector representation from a high-dimensional input sentence, and decoding generates a correct high-dimensional target translation from that representation.

\item[$\bullet$] \emph{SA module}: The SA module is used to speed up training by partially replacing expensive ground-truth fitness evaluations. The surrogate model uses all individuals in the population and their corresponding true fitness values as historical data to train the model, and then uses the trained model to evaluate candidate solutions and select the optimal solution ${y^*}$ from them.

\end{itemize}

\subsection{PE module : Random Embedding}
\label{sect:PE}

Random Embedding refers to a dimensional reduction technique with technically simple and desirable theoretical property, which project data into low-dimensional spaces with a randomly generated matrix \cite{RE_1,RE_Bayesian,NCSRE}. The implicit assumption is that, for the SAERL framework, the fitness of candidate solutions are only affected by a few dimensions instead of all. This assumption is technically reasonable for the DNN-based policy since DNN often preserves redundent weights and can be effectively prunned \cite{Gui_purnning_1,Gui_purning_3,hong_pruning_1}.

Based on this assumption, given a high-dimensional function with low ${S}$-efficient dimension (i.e., ${d_e \ll D}$) and a random embedding matrix ${A \in \mathbb{R}^{D \times d}}$, for any ${x \in \mathbb{R}^{D}}$, there must exist ${y \in \mathbb{R}^{d} }$ such that ${f(Ay) = f(x)}$ with probability 1. That is, random embedding enables us to optimize the lower-dimensional function ${g(y) = f(Ay)}$ in ${\mathbb{R}^{d}}$ instead of optimizing the original high-dimensional ${f(x)}$ in ${\mathbb{R}^{D}}$, while the function value is still evaluated in the original solution space. For ${S}$-efficient dimension definition and provable completeness guarantee, please refer to \cite{RE_1}.

Specifically, at encoding steps, a random embedding matrix ${A \in \mathbb{R}^{D \times d}}$ is generated drawn from ${\mathcal{N}(0, 1)}$. For each solution ${x}$, there must be a unique ${y}$ corresponding to it, according to ${x=Ay}$. On the embedding space, the surrogate preselects the best solution from ${M}$ candidate solutions. 
The final candidate is then decoded to the original high-dimensional solution space.

\subsection{PE-FCPS-NCS: a concrete algorithm}
\label{sect:PE-FCPS-NCS}

As mentioned above, the PE module provides the possibility for SA to be used in ERL. To instantiated the PE-SAERL for empirical studies,  we choose Negatively Correlated Search (NCS) \cite{NCS} as the EA module and the Fuzzy Classification Pre-Selection (FCPS) \cite{zhang_FCPS} as the surrogate module. The pseudo-code for the concrete algorithm is given in Algorithm \ref{alg_pe_fcps_ncs}.



\begin{algorithm}[htb]
    \renewcommand{\algorithmicrequire}{\textbf{Input:}}
    \renewcommand{\algorithmicensure}{\textbf{Output:}}
    \caption{PE-FCPS-NCS}
    \label{alg_pe_fcps_ncs}
    \begin{algorithmic}[1]
        \REQUIRE Number of sub-process ${\lambda}$; RL simulator $f$; ${S}$-Effective embedding dimension $d$; Origin policy dimension $D$; Evaluation limitation $max\_steps$; Number of candidate policies ${M}$.
        \ENSURE BestFound policy ${x}^{*}$.
        \STATE Initialize ${\lambda}$ policies ${{x}_{i}} \in \mathcal{R}^D$ uniformly,  ${i=1,\cdots,\lambda}$.
        \STATE Evaluate the $\lambda$ policies with respect to the objective function $f$.
        \REPEAT
            \FOR{$i=1$ to $\lambda$}
                \STATE Sample candidate policies $\{{x}_{i,1}^{\prime},\cdots,x_{i,M}^{\prime}\}$ from the distribution ${p_{i} \sim \mathcal{N}(x_i,\Sigma_i)}$; 
                \STATE Generate a random matrices ${A \in \mathcal{R}^{D \times d}}$ with ${\mathcal{N}(0, I)}$;
                \STATE Encoding with ${y_{i,j}=A^{-1}\cdot{x}_{i,j}^{\prime}}$ where ${j=1,\cdots,M}$;
                \STATE Choose a policy ${y_i^*}$ with maximal membership degree belongs to $"$promising$"$ class by a fuzzy classifier model ${m}$;
                \STATE Let ${{x}_{i}^{\prime}=A \cdot y_i^*}$;
                \STATE Evaluate ${f({x}_{i}^{\prime})}$ as the qualities of ${{x}_{i}^{\prime}}$;
                \STATE Calculate $d(p_{i})$ and $d(p_{i}^{\prime})$, where $p_{i}^{\prime} \sim \mathcal{N}({x}_{i}^{\prime}, {\Sigma}_{i})$;
                \IF{${f({x}_{i}^{\prime})+\varphi\cdot d(p_{i}^{\prime})>f({x}_{i})+\varphi \cdot d(p_{i})}$} 
                    \STATE ${x}_{i}={x}_{i}^{\prime}$; and $p_{i}=p_{i}^{\prime}$;
                \ENDIF
                \STATE Update ${\Sigma}_{i}$ according to the $1/5$ successful rule;
            \ENDFOR
        \UNTIL{$steps\_passed \geq  max\_steps$}
    \end{algorithmic}
\end{algorithm}

Some components in Algorithm \ref{alg_pe_fcps_ncs} are explained as follows.
\begin{itemize}
\item[*] \emph{Initialization}: A set of policies ${\{x_1,x_2,\dots,x_\lambda\}}$ are initialized uniformly and first evaluated with respect to the RL simulator ${f}$ in lines ${1 \sim 2}$.
\item[*] \emph{FCPS section}: In Line ${5}$, ${M}$ candidate policies are sampled by means of a Gaussian operator. In Line ${8}$, the high-rank policies with the maximal membership degree belongs to "promising" class are chosen out. And one is randomly selected from the "promising" high-rank policies as the candidate ${y^{*}}$.
\item[*] \emph{PT section}: A Gaussian random matrix is generated in line ${6}$ and applied in line ${7}$ to encode the candidate policies from the original high-dimensional policy space into a low-dimensional effective subspace. Then, the policy space is decoded in line ${9}$.
\item[*] \emph{NCS section}: Considering both the quality and the diversity, i.e., ${f({x}_{i}^{\prime})+\varphi\cdot d(p_{i}^{\prime})}$ and ${f({x}_{i})+\varphi \cdot d(p_{i})}$, the following steps are carried out. In line ${10}$, the candidate policy is evaluated with respect to ${f}$. And the diversity value of the current and candidate policies are also calculated as a basis for environmental selection in line ${11}$. In Lines ${12 \sim 14}$, the better one between the current and candidate policy is selected into the next generation. Finally, the step size is updated as described in line ${15}$.
\item[*] \emph{Stopping condition}: In Line ${17}$, when the time budget has run out, i.e., the number of times the policy interacts with the environment exceeds the given maximum number ${max\_steps}$, the best solution ${x^{*}}$ ever found is output.
\end{itemize}

\section{Experiments}
\label{sect:experiments}

In this experiment, we aim to answer the following questions: (1) Does the surrogate model really improve ERL compared to baselines? (2) Does the surrogate model really speed up training compared to original NCS? (3) Is the proposed method sensitive to parameters?

\subsection{Experiments settings}
\label{sect:experiments_settings}

\subsubsection{Environment}
We apply the algorithm to a series of Atari 2600 games implemented in The Arcade Learning Environment (ALE) \cite{ALE}. The Atari 2600 is a challenging RL testbed that provides agents with high-dimensional visual input and a diverse and interesting set of tasks, including obstacle avoidance (e.g. Freeway), shooting (e.g. Beamrider), two-player (e.g. DoubleDunk) and other types. These types of games provide agents with significantly different environmental settings and ways of interacting, thus enabling testing of RL methods with different tasks in maximizing long-term rewards. All these tasks are packaged according to the standard OpenAI Gym API \cite{gym} and are simulated friendly. 

\subsubsection{Baselines}

Four state-of-the-art algorithms, namely A3C \cite{A3C}, PPO \cite{PPO}, CES \cite{ERL_CES}, and NCS \cite{NCS}, are used as the major baselines and follow all original hyperparameter settings. Among them, PPO and A3C are popular gradient-based methods that use traditional backpropagation to train networks. CES is a recently proposed ERL that utilizes a canonical evolution strategy to directly optimize the weights of policy neural networks. Besides, to show how PE-SA promotes NCS, we directly apply NCS as a baseline to Atari games.

\subsubsection{Performance metric}

The quality of the policy is measured by the testing score, i.e., the average score over 30 repetitions of the game without the frame limitations. 
More specifically, the testing score refers to the cumulative reward of the policy in an episode, and the episode refers to the time when there is no action in the random "noop" frame (sample in the interval 0 to 30) at the beginning of playing until the end signal is received, as mentioned by Machado et al. \cite{ALE}.
Technically, to prevent the agent from falling into a "dead situation", the frames of one episode is limited to a very large value (i.e. 100,000 frames).

\subsubsection{Other protocols}

All the baselines share the policy network.We directly adopt the policy network proposed by Mnih et al. \cite{DQN}, which consists of three convolutional layers and two fully connected layers. 
The network architecture is shown in Table \ref{table-policy-network-architecture}, which involves nearly 1.7 million connection weights that need to be optimized. 
As suggested by Mnih et al. \cite{DQN}, the raw observations are resized to 84x84, and the skipping frame is set to 4. 
The time budget is set to the consumed total game frames allowed in each training phase. 
For ERL methods (i.e., CES, NCS), The total game frames are set to 0.1B. 
For gradient-based methods (i.e., PPO), the time budget is set to 0.04B for fairness, as it works via back-propagation and the ratio of consumed frames on the same CPU-cored hardware is 2.5 \cite{CCNCS}. 
Table \ref{table-PE-FCPS-NCS-para} shows the hyper-parameters of PE-FCPS-NCS in this work, which are common across all environments. 

\begin{table*}   
\begin{center}   
\caption{The network architecture of the policy.}  
\label{table-policy-network-architecture}
\begin{tabular}{lllllll}
\hline & Input size & Output size & Kernel size & Stride & \#filters & activation \\
\hline Conv1 & $4 \times 84 \times 84$ & $32 \times 20 \times 20$ & $8 \times 8$ & 4 & 32 & ReLU \\
Conv2 & $32 \times 20 \times 204$ & $64 \times 9 \times 9$ & $4 \times 4$ & 2 & 64 & ReLU \\
Conv3 & $64 \times 9 \times 9$ & $64 \times 7 \times 7$ & $3 \times 3$ & 1 & 64 & ReLU \\
Fc1 & $64 \times 7 \times 7$ & 512 & $-$ & $-$ & $-$ & ReLU \\
Fc2 & 512 & \#Action & $-$ & $-$ & $-$ & $-$ \\
\hline
\end{tabular}
\end{center}   
\end{table*}

\begin{table*}  
\begin{center}   
\caption{The major hyperparameter settings of PE-FCPS-NCS. The hyperparameters consists of three parts : NCS \cite{NCS}, FCPS \cite{zhang_FCPS}, and random embedding \cite{RE_1}.}  
\label{table-PE-FCPS-NCS-para}
\begin{tabular}{lll}
\hline 
\textbf{Parameter} & \textbf{Value} & \textbf{Remark} \\
\hline 
$\lambda$ & 6 & The number of processes, that is, the number of populations \\
$d$ & 100 &  Effective subspace dimension of random embedding\\
$epoch$ & 5 & The update interval of covariance matrix \\
$r$ & 1.2 &  Learning rate for covariance matrix \\
$M$ & 3 &  The number of candidate solutions \\
${\varphi}$ & 1.0 &  The trade-off factor between quality and correlation\\
$[l,h]$ & [-0.1,0.1] &  Search space boundaries in low-dimensional spaces \\
\hline
\end{tabular}
\end{center}   
\end{table*}


\subsection{Results and analysis}
\label{sect:results}

\subsubsection{Comparisons with baselines} 

For each game, we train a neural network policy model and do 30 repeated testings of the trained model to obtain a average testing score as performance. To eliminate bias, we repeat the process three times, and the testing scores of each algorithm on 5 games are shown in Table \ref{table-result-main}. 
The Average row shows the average performance of the three executions where the best performance is marked in bold, and the Percent row indicates the average performance as a percentage of the best performance. 
It can be seen that PE-FCPS-NCS generally performs the best among all compared algorithms. In particular, in the Freeway and Bowling games, the best performances of baselines are only 65.23\% and 69.49\% compared to our proposed method. 
In the two-player game (DoubleDunk), a high score with a positive score is preferred, and a negative score implies agent failure. 
However, all algorithms score are negative, but relatively speaking, our algorithm performed best. 
Note that in the Alien game, PE-FCPS-NCS, while better than traditional gradient-based reinforcement learning algorithms (A3C, PPO) and evolution-based algorithms (CES), performs worse than the original NCS algorithm. 
In some specific tasks, the application of surrogate may encourage exploration in a worse direction.

\begin{table*}   
\begin{center}   
\caption{The performance of PE-FCPS-NCS and four baselines on 5 Atari games. Priority is given to higher average scores in all games.}  
\label{table-result-main}
\begin{tabular}{lllllll}
\hline 
\textbf{Game} & \textbf{Performance} & \textbf{A3C} & \textbf{PPO} & \textbf{CES} & \textbf{NCS} & \textbf{PE-FCPS-NCS} \\
\hline 
\textbf{Time Budget} & & 40M & 40M & 0.1B & 0.1B & 0.1B\\
\textbf{Alien} 
& Average & 766.00 & 638.20 & 561.80 & \textbf{1043.40} & 825.68  \\
& Percent & 73.41\% & 61.17\% & 53.84\% & 100.00\% & 79.13\% \\
\textbf{BeamRider} 
& Average & 633.80 & 600.80 & 433.10 & 703.80 & \textbf{724.80}  \\
& Percent & 87.44\% & 82.89\% & 59.75\% & 97.10\% & 100.00\%  \\
\textbf{Bowling} 
& Average & 0 & 23.40 & 25 & 64.53 & \textbf{92.87}  \\
& Percent & 0.00\% & 25.20\% & 26.92\% & 69.49\% & 100.00\%  \\
\textbf{DoubleDunk} 
& Average & -2 & -3.1 & -3.7 & -1.3 & \textbf{-0.60} \\
& Percent & 54.78\% & 19.33\% & 0.00\% & 77.34\% & 100.00\%  \\
\textbf{Freeway}
& Average & 0.00 & 14.8 & 14.2 & 16.7 & \textbf{22.69}  \\
& Percent & 0.00\% & 65.23\% & 62.58\% & 73.60\% & 100.00\%  \\
\hline
\end{tabular}
\end{center}   
\end{table*}

\subsubsection{Computational time analysis}

\begin{figure}
\centering
\includegraphics[width=0.64\columnwidth]{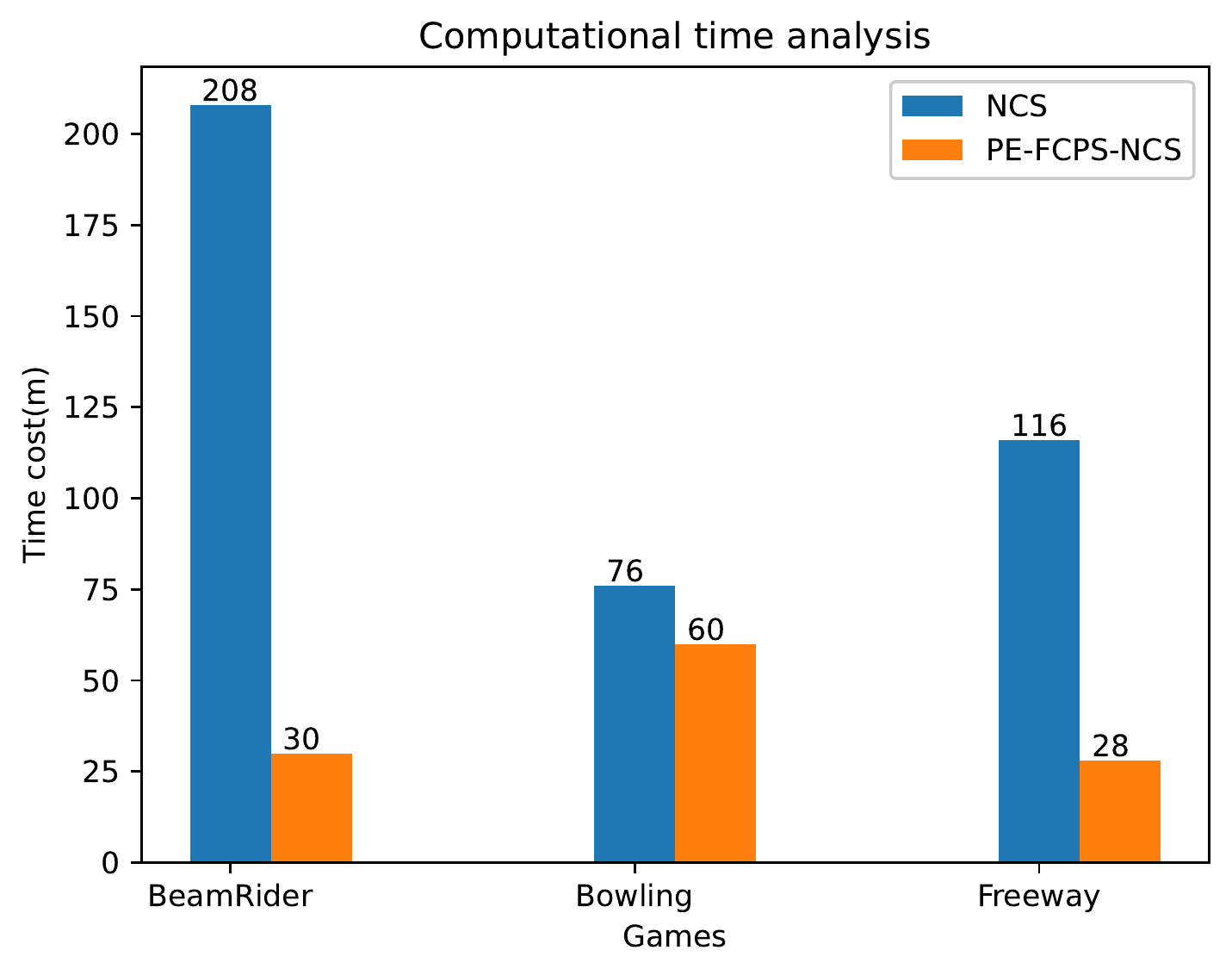} 
\caption{The training time costs (in minutes) of NCS and PE-FCPS-NCS on three games (i.e., BeamRider, Bowling, Freeway) to reach a specified score that achieved by the NCS method after training for 10 million game frames. }
\label{fig_time}
\end{figure}

Intuitively, we need to verify whether the surrogate model can speed up training. More specifically, we compare the training time consumed by NCS and PE-FCPS-NCS to reach a specified score. This score is set to the test performance achieved by the NCS method after training for 10 million game frames. The computational time for the PE-FCPS-NCS method to reach that score is then counted. Fig. \ref{fig_time} shows the computational time costs (in minutes) of NCS and PE-FCPS-NCS on three games (i.e., BeamRider, Bowling, Freeway). 
The test performance achieved by the NCS method is 656, 34 and 14.9, respectively.
From Fig. \ref{fig_time}, it can be seen that the computational time cost of the surrogate-assisted NCS is quite superior to the original NCS. In particular, in the BeamRider game, the time consumption of the original NCS is nearly 7 times that of PE-FCPS-NCS. In addition, the insignificant difference in the Bowling game is mainly because the number of game frames consumed by each iteration is extremely large, which is 
nearly 4 times that of Freeway. Considering that some additional computational burden is introduced, e.g., the training of the surrogate model and the computation of the embedding, it suffices to show that PE-FCPS-NCS is promising not only for its solution performance but also in accelerated training.

\subsubsection{Sensitivity analysis}

\begin{figure}
\centering
\includegraphics[width=0.64\columnwidth]{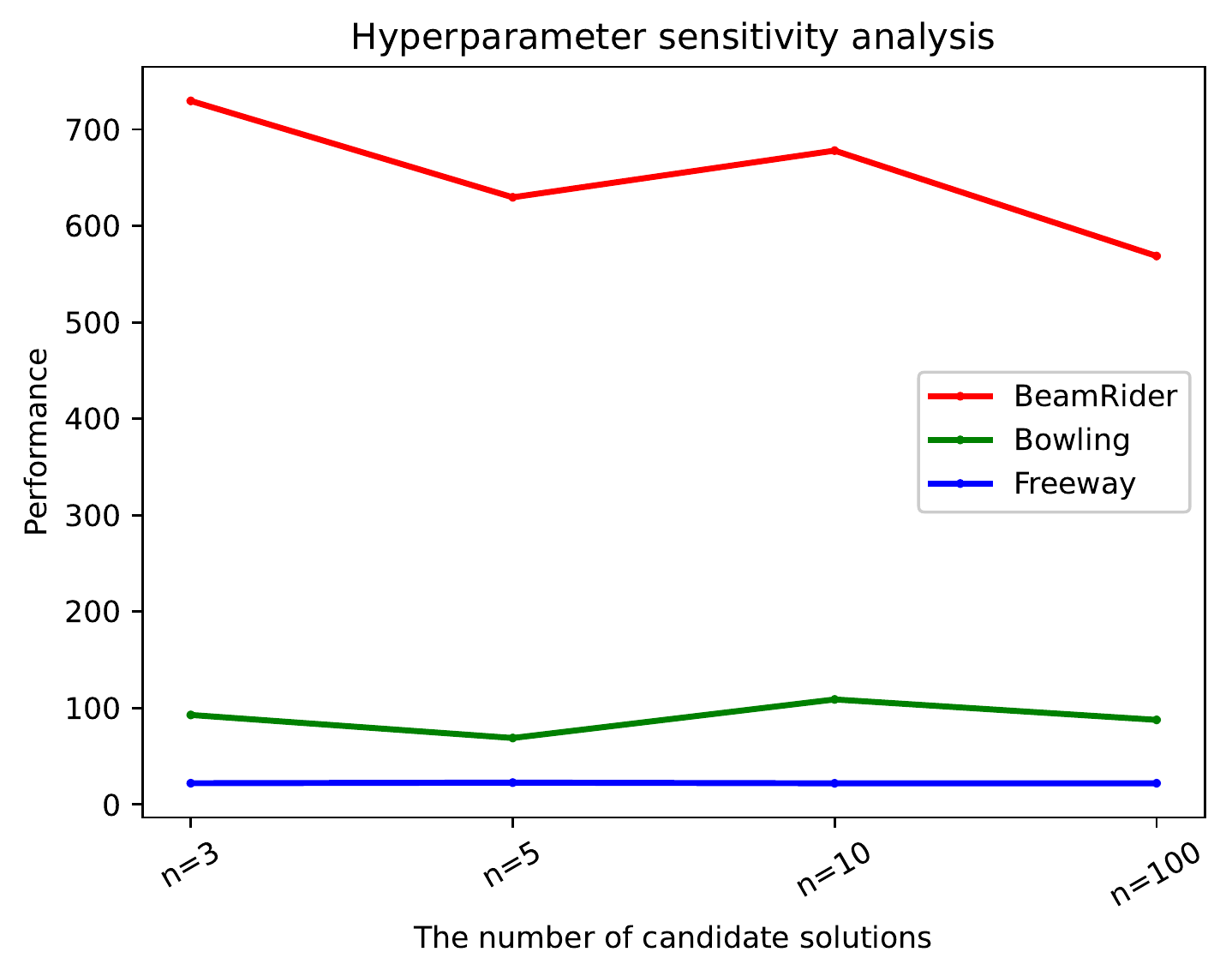} 
\caption{The performance curves as the number of candidate policies vary in [3, 5, 10, 100].}
\label{fig_sensitivity}
\end{figure}

We vary the parameter value over a wide range and re-evaluate to perform parameter sensitivity analysis. More specifically, we choose the number of candidate policies involved in pre-selection in the surrogate model as the hyperparameter, and set it to four different values as [3, 5, 10, 100]. We also select three games (i.e., BeamRider, Bowling, Freeway) for sensitivity analysis. As shown in Fig. \ref{fig_sensitivity}, the performance curves did not change drastically with the perturbation of the parameter, which means that PE-FCPS-NCS is usually not very sensitive to this important parameter.

\section{Conclusion}
\label{sect:conclusion}

This paper studies how to effectively enable surrogate-assisted evolutionary reinforcement learning to speed up training. First, To apply surrogate to preselect millions of connection weights of neural network policies, this paper employs the PE-SAERL Framework to scale up surrogate for large-scale search space. Next, we propose a concrete algorithm based on the framework, which called PE-FCPS-NCS. Then, empirical studies are conducted on 5 Atari games to verify the proposed method. Empirical results show that the proposed method can perform more efficiently than the four state-of-the-art algorithms (i.e., PPO, A3C, CES, NCS), while effectively accelerating training compared to the original NCS. Last, this paper studies the parameters sensitivity of the proposed method.

\label{sect:bib}
\bibliographystyle{splncs04}
\bibliography{references}






\end{document}